# Transformer-Based Deep Learning Model for Bored Pile Load-Deformation Prediction in Bangkok Subsoil

Sompote Youwai and Chissanupong Thongnoo

*Abstract*— This paper presents a novel deep learning model based on the transformer architecture to predict the load-deformation behavior of large bored piles in Bangkok subsoil. The model encodes the soil profile and pile features as tokenization input, and generates the load-deformation curve as output. The model also incorporates the previous sequential data of load-deformation curve into the decoder to improve the prediction accuracy. The model also incorporates the previous sequential data of load-deformation curve into the decoder. The model shows a satisfactory accuracy and generalization ability for the load-deformation curve prediction, with a mean absolute error of 5.72% for the test data. The model could also be used for parametric analysis and design optimization of piles under different soil and pile conditions, pile cross section, pile length and type of pile.

## I. INTRODUCTION

Deep learning (DL) is a branch of artificial intelligence that aims to learn from data and generate new outputs [1, 2]. DL started with solving regression problems [3], where the goal was to predict a single value from multiple inputs. Later, DL expanded to other tasks, such as understanding the meaning of text or images. One of the popular DL models for image classification is the convolutional neural network (CNN), which can recognize different types of images by extracting features from them[4]. Recently, DL has reached the stage of 'generative modeling,' where the goal is to create new data from existing data. This development is accelerated by the introduction of the 'Transformer'[5], a novel neural network architecture that can handle sequential data with attention mechanisms. Various domains, such as text-to-text [6–8], text-to-image [9] and image to text generation [10, 11] use the Transformer architecture as a powerful and effective deep learning approach for generative modelling. The main innovation of this architecture is the use of attention mechanisms [5], which allow a neural network to concentrate on relevant parts of the input or output depending on the task. This technique resembles the human attention process, which eliminates irrelevant information and enhances the learning of important information. Attention mechanisms are the foundation for large-scale pre-trained language models, such GPT-4[7], BERT [6], and LLaMa [8] which have achieved outstanding results in natural language processing tasks. The Transformer architecture comprises an encoder and a decoder, which are connected by latent variables. The Transformer model is a neural network architecture that uses attention mechanisms to encode and decode natural language sentences

Prediction of pile behavior in complex soil conditions is essential for the optimal design and construction of deep foundation systems. The subsoil of Bangkok presents significant geotechnical challenges due to the presence of alternating clay and sand layers with high spatial variability and compressibility. These complexities require advanced computational techniques to simulate pile responses under different loading scenarios accurately. Conventionally, pile deformation in Bangkok's subsoil is verified by expensive field pile loading tests, either static or dynamic, to ensure the pile capacity and settlement. Consequently, a large amount of pile load test data is available for Bangkok's subsoil. However, the cost per pile loading test in Bangkok is very high, ranging from 17,000 USD to 28,000 USD for large rectangular piles. Therefore, developing a novel method to predict pile behavior under increasing loading conditions is beneficial. Several methods have been proposed to predict pile behavior in Bangkok's subsoil, including analytical approaches, Finite Element Analysis (FEA) [12, 13] [14]. The finite element analysis requires a sophisticated constitutive model, interpretation of a model parameter and complicated techniques to conduct the analysis. Therefore, there is a need for a tool that can predict the unforeseen pile behaviors using only the input of the soil testing results from borehole, SPT value and soil type. The novel approach should be able to generate the load-deformation curve at different loading levels, different pile lengths and dimensions (diameter or width of pile) and soil profile.

Several studies have adopted DL techniques to investigate pile behaviors, such as load-settlement curve prediction, dynamic pile loading test analysis, and soil-pile interaction modeling [15, 16, 16–20]. However, the predicting results were not satisfactory with high error and trend o be overfitting. The most of these studies have used fully connected neural networks (FNN) as the main DL architecture, which take pile properties (e.g., diameter and length) as input parameters. Some studies have also incorporated soil profiles [16], pile configuration and load increment characteristics as additional features to train the FCNNs. A limitation of the FNN is that they do not consider the sequential data of soil profile or the previous time step of the load deformation curve, which are important factors for the

Sompote Youwai, Associate Professor, Department of Civil Engineering, King Mongkut's University of technology Thonburi
Chissanupong Thongnoo, Master student, Department of Civil Engineering, King Mongkut's University of technology Thonburi

behavior of pile deformation. Therefore, there is a need for more advanced DL models that can capture the temporal and spatial dependencies of pile-related phenomena handle complex tasks involving both sequential and spatial data. We hypothesize that the architecture of transformer based on large language mode could be adopted to apply in prediction of load-deformation of pile. In this study, we propose to apply the Transformer model to a regression problem in the domain of discrete soil property information, such as borehole data or field tests like Cone Penetration Testing (CPT). We embed engineering-specific properties into the tensors, which are vectors that represent the input token and their positional embeddings. The encoder applies an attention-based correlation process to generate query responses, which are then used by the decoder to produce the output tokens. We also incorporate the previous sequential data of load-deformation curve of pile into the decoder, similar to the architecture of the language translation model. To the best of our knowledge, there is scarce research on using Transformer-based generative models as predictive models in geotechnical engineering. The trained deep learning model can be used as a generative model similar to large language models. By inputting the initial soil profile and feature, the generative model can predict desired geotechnical outcomes. This might be beneficial for geotechnical projects in design and supervision to predict the unforeseen value prior and during construction.

This study proposes a conditional transformer-based model (CTM) for simulating the load-deformation of piles in Bangkok subsoil. The CTM can account for different conditions or features of piles, such as length, dimension, and type, by using a transformer model. The study consists of four steps: (1) applying feature engineering to characterize the soil profile and the data of various pile types, (2) encoding the site-specific geotechnical engineering data into the CTM, (3) decoding the encoded latent variables to predict the pile load for a given deformation, incorporating the pile condition into the decoding process, and (4) optimizing the transformer model hyperparameters to minimize the prediction error. The study also illustrates how the trained CTM can be used to predict the load-deformation of piles and conduct a parametric analysis. The trained CTM is publicly available on GitHub for geotechnical engineers to simulate pile-load deformation in subsoil models. The CTM is implemented using open-source deep learning libraries [21] and the Python programming language [22], ensuring its adaptability and scalability for predicting pile behaviors across different soil conditions and geographical locations worldwide. This study contributes to the field of geotechnical engineering by providing a valuable tool for accurately predicting pile behaviors in both Bangkok's subsoil and other regions.

## 2. RELATED WORK

Neural networks have been widely used to predict the pile behaviour in the last decade, as shown in Table 1. Most of the previous studies applied fully connected neural network (FNN) to predict the load-deformation curve of pile. [23] first introduced the concept of sequence to sequence model for this task, treating the pile load as a sequential data corresponding to certain deformation. The prediction of the next load step used the previous load sequence as the output for neural network. The predicted sequential load matched well with the actual testing data. [15] used the results of cone penetration test as an input data to predict the settlement of pile. They also added the pile features such as length and rigidity as input features for neural network. However, the root mean square error (RMSE) was still high. The soil input data was the average standard penetration test data with five intervals along pile shaft. [17, 20] proposed a hybrid model between FNN and other methods, such as random forest and optimization method, to predict the result of pile load test. However, their model also suffered from high error and poor generalization to unseen test data. In this research, we applied a sequence-to-sequence neural network model to predict the pile deformation, using sequential data such as soil profile and load-deformation curve of piles as input data. The proposed model was based on the transformer structure, which will be discussed in the next section.

Table 1 Related work to this study

| Model | Paper | Dataset | Predicted value | Remarks |
|---|---|---|---|---|
| FNN hybrid with optimization | [20] | 50 dynamic pile load tests | Pile end bearing | |
| FNN | [16] | 3 cases in UAE | Load-deformation | High different between validation data and trained |
| FNN | [19] | Pile load test | Pile capacity | |
| FNN, RDF | [17] | 2314 Pile load test | Pile capacity | High error |
| FNN | [15] | 1013 pile load tests | Settlement | Apply CPT value as input |
| FNN | [23] | 23 pile load tests | Sequential of load | Sequence to Sequence struc |
| FNN | Fully connected neural network | | | |
| RDD | Random forest | | | |

## 3. BACKGROUND

The sequence to sequence (seq2seq) model [24] is a general framework for sequence learning that uses a multilayered Long Short-Term Memory (LSTM) network to map the input sequence into a fixed-length vector, and then another LSTM network to generate the target sequence from the vector. This model was first proposed for language translation tasks, such as English to French translation. The transformer [5] is an extension of the seq2seq model that replaces the recurrent layers with self-attention layers (Fig. 1). The transformer consists of an encoder and a decoder, each composed of multiple stacked self-attention and feed-forward layers. The encoder encodes the input text into a sequence of context-aware vectors, which are then passed to the decoder through a cross-attention layer. To facilitate the training of deep neural networks, residual connections are applied to add the input of a layer to its output, creating a shortcut path for the information. This way, the network can learn to preserve the important features of the input and avoid losing them due to non-linear transformations or vanishing gradients. Layer normalization is also used to normalize the inputs of each layer across the features dimension, which is the last dimension in the input tensor. Layer normalization helps to stabilize the training process and improve the generalization performance of transformer models. The decoder produces the output text by attending to both the encoder output and its own previous output. The transformer model has achieved state-of-the-art results on various natural language processing tasks, such as machine translation, text summarization, and natural language generation.

The attention mechanism is a technique that enables a neural network to selectively attend to the most relevant parts of the input or output sequence, depending on the task. It is widely used in natural language processing (NLP) to enhance the performance of encoder-decoder models, such as machine translation, text summarization, and speech recognition. The attention mechanism operates by projecting the input vector onto a trainable vector at the attention head, Q, K, and V (Fig. 2). The vectors at the attention head, which are query, key and value, can be learned during the neural network training. To calculate the query, key and value vectors, q, k and v, the attention mechanism can be viewed as a function that takes a query vector and a set of key-value pairs as inputs, and returns a weighted sum of the value vectors as output. The query vector represents the current state of the decoder, and the key-value pairs represent the encoded input sequence. The weights are computed by measuring the similarity between the query vector and each key vector, using a scoring function such as dot product, cosine similarity, or a neural network. The weights are then divided by the square root of the key dimension to reduce the effects of high sequence dimension (Eq.6). The weights are then normalized by a softmax function to form a probability distribution. The result from the softmax function is then multiplied with the value vector to obtain the output. The output from the attention mechanism is then added with the initial input vector, known as residual connection, to mitigate the effects of gradient descent.

$$Attention(q, k, v) = softmax\left(\frac{qk^T}{\sqrt{d_k}}\right)v \qquad (1)$$

## 4 TRANSFORMER MODEL

We adopted the basic model architecture of the transformer from the original paper by Vaswani et al. (2017) for this research (Fig. 3). The model consists of three parts: an encoder, a decoder, and a multilayer perceptron (MLP). The encoder takes the soil profile as input, the decoder generates the loading output from pile features, the output shift right, and the MLP produces the final output. The model aims to predict the next loading for a given applied deformation to the pile. The decoder input includes the different pile conditions (features), which are concatenated with the latent variable from the encoder and the previous sequential data of the load-settlement curve of the pile. The MLP then uses this input to predict the next loading step.

### 4.1 The input tensor

The encoder input tensor consists of the soil borehole data and different pile features (Fig. 4). Based on the success of the transformer model in language and image classification tasks, we hypothesize that the model can be adapted to geotechnical engineering applications. The initial soil profile can be represented by a sequence of tokens, where each token is a one-dimensional vector that encodes its meaning. Different tokens may have similar vectors if they have similar meanings. The geotechnical data for each depth can be represented by a vector with its strength, soil type, or other relevant features. Each depth can be considered as a token in a large language model. This study uses the standard penetration value (SPT) as a strength indicator of soil. If a certain layer does not have an SPT value, such as when tested by the unconfined compression test method, the undrained shear strength is converted to SPT (N) value [25] using the following equation:

$$N = \frac{S_u}{5} \qquad (2)$$

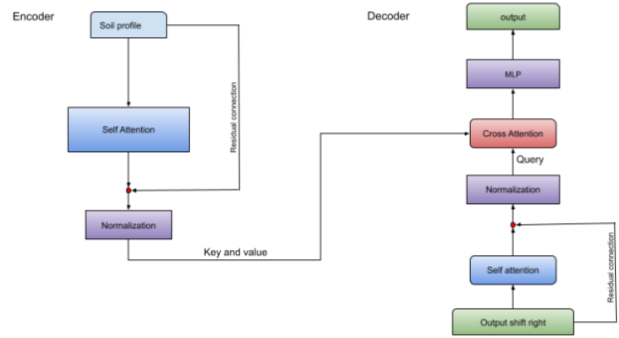

Figure 1 The architecture of transformer

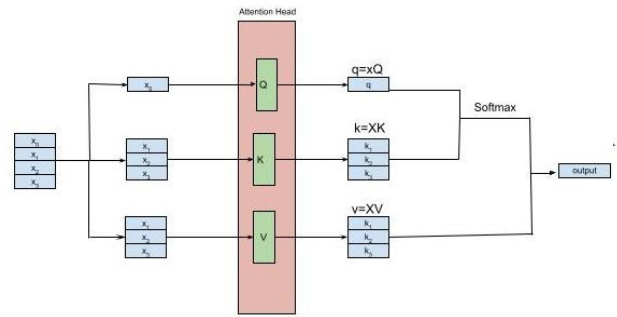

Figure 2 The diagram of attention mechanism

The vector consists of several components that describe the soil and pile characteristics. The first component is the soil type, which is encoded as an integer: 1 for clay and 2 for sand layer. The second component (pile in layer) is a binary indicator of the presence of the pile in the soil layer: 1 if the pile exists, and 0 otherwise. The third and fourth components are the width and height of the pile, respectively. For a circular pile, these are equal to the diameter of the circle. The fifth component is the type of pile, which is also encoded as an integer: 1 for a circular bored pile, and 2 for a barrette pile. These two types of piles are distinguished because they have different behaviors due to their construction techniques. The friction and end bearing resistance of a barrette pile are usually lower than those of a circular bored pile. The vector has 60 columns, each representing a depth of soil layer in meters. The input tensor is constructed from the ground surface to a depth of 60 m. If there are no available testing results of borehole below 60 m, the last value of SPT and soil type are used as the properties of the missing soil layers.

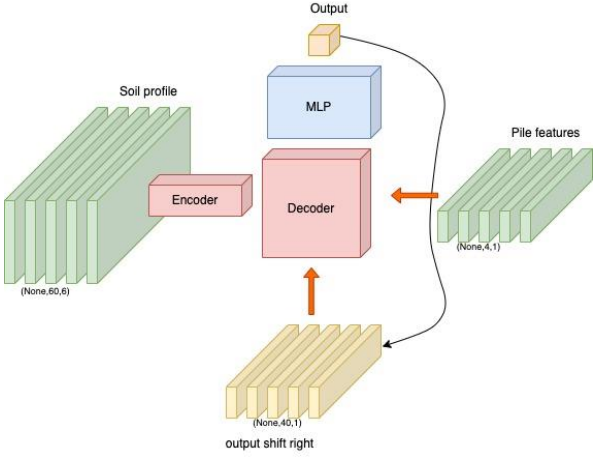

Figure 3 The architecture of transformer model for prediction pile deformation

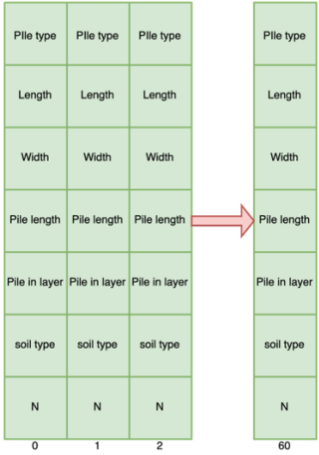

Figure 4 The input vector for encoding

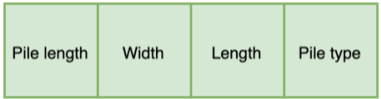

Figure 5 Conditional vector for decoder

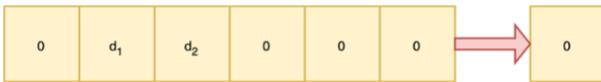

Figure 6 Output shift right

The decoder input consists of two parts: a conditional vector and an output shifted right, as shown in Figs. 5 and 6. The conditional vector for each training data has four columns (Fig. 5), which are pile length, width of pile, length of the pile cross section, and pile type. The integer encoder for categorical data is similar to the previous section. The output for predicting the load-deformation of pile is a sequence of loads corresponding to the deformation increments of 1 mm. Fig. 6 shows an example of output shifted right, which represents the prediction for the fourth loading step. The initial value is added with zero to shift the results to the right. The first and second loading steps are input to the tensor to predict the third loading step. The other values that do not have load values are padded with zero to prevent the attention from using them in the calculation. The dimension of the sequential loading results is 40 columns, which corresponds to the maximum deformation of 40 mm. If the load deformation data does not contain the deformation up to 40 mm, the initial input results are set as -1. The samples that have -1 values in the sequence are removed from the training data.

*4.2 Model architecture*

The encoder extracts features or patterns from the input vector, which represents the tokenized soil profile data (Fig 7). The input tensor undergoes a multi-head attention layer, implemented with the Keras library [26]. Multi-head attention splits the query, key, and value vectors into multiple sub-vectors of smaller dimension. Each sub-vector enters a scaled dot-product attention function, which calculates the attention weights and the output vector. The output vectors concatenate and project to the final output dimension. The multi-head attention equation is:

$$\text{MultiHead}(q,k,v) = \text{Concat}(\text{head}_1, \ldots, \text{head}_h)W^O \quad (3)$$

$$head_i = Attention(qW_i^Q, kW_i^K, vW_i^V) \quad (4)$$

$$\text{Attention}(q,k,v) = \text{softmax}(\frac{qk^T}{\sqrt{d_k}})v \quad (5)$$

Next, the output vector is normalized to reduce overfitting. Normalization is a technique that transforms the input data into a standard distribution, typically with zero mean and unit variance. Normalization can enhance the performance and stability of neural networks. The equation for normalization is:

$$x_{normalized} = \frac{x-\mu}{\sqrt{\sigma^2+\epsilon}} \quad (6)$$

Using a residual connection technique, the vector was added to the previous value of the input vector. A residual connection allows deep neural networks to learn more effectively by adding the information from previous layers to the current layer's output. This helps the network preserve the essential features of the input and avoid the problems of vanishing or exploding gradients. The above process was iterated in a loop, with the number of iterations being a hyperparameter of the model. Too many iterations might make the model overly complex, hard to train, or prone to collapse. Finally, the tensor was flattened to a one-dimensional latent variable and passed to the decoder part of the model.

The decoding part of the architecture is illustrated in Fig. 8. The output shift right tensor was first input to the multi-head attention layer. The dropout layer acted as a regularization technique to prevent overfitting. Then, the tensor was added to the normalization and residual connection layer with a specified number of iterations. Next, the output was flattened to have the same dimension as the latent variable. The three components of the vector, namely the latent variable, the output from the output shift right, and the pile feature, were concatenated into a single vector. This vector was then fed into a multilayer perceptron layer (MLP), which consisted of conventional fully connected neurons (dense). The number of neurons was gradually reduced to a single output value, which represented the prediction of the next sequential load.

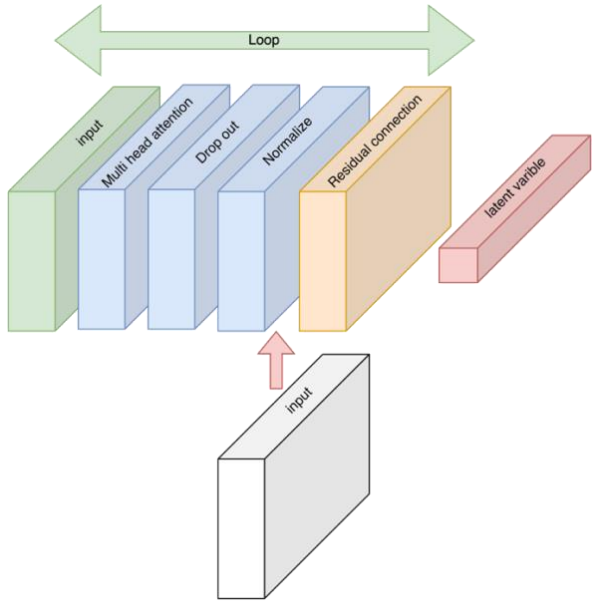

Figure. 7 Model architecture for encoder

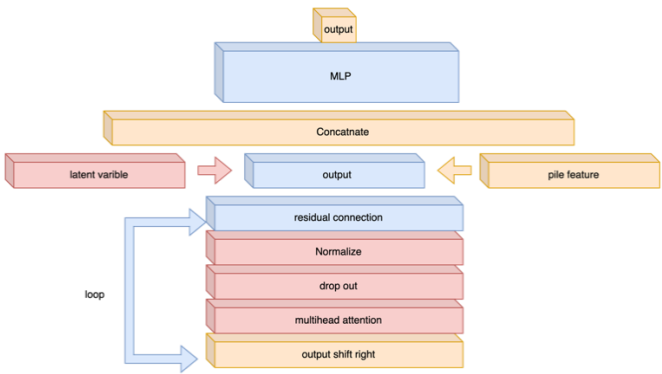

Figure 8 Model architecture for decoder

## 5. DATA CHARACTERISTICS

A model was trained to predict the load-deformation behavior of bored piles based on the results of static pile load tests conducted in the Bangkok area. The bored piles had either circular or rectangular cross-sections, and the latter were referred to as barrettes. A total of 58 test data sets were used for model training. The circular piles had diameters ranging from 0.8 to 1.8 m, while the barrettes had widths of 1 or 1.2 m and cross section lengths of 3 or 3.8 m. The pile sizes followed a distribution that showed most piles had diameters between 0.5 and 1.5 m (Fig. 9). Eight barrettes were included in the model training. The pile lengths varied from 39 to 60 m (Fig. 10), depending on the intention to place the pile tip on sand or stiff clay layer, which affected the end bearing and settlement characteristics. The pile length, dimension, and type were encoded as a vector for training in the encoder part (Fig. 4). The presence or absence of pile in each soil layer was also encoded as a binary value in the row of pile in layer in the token of each soil layer.

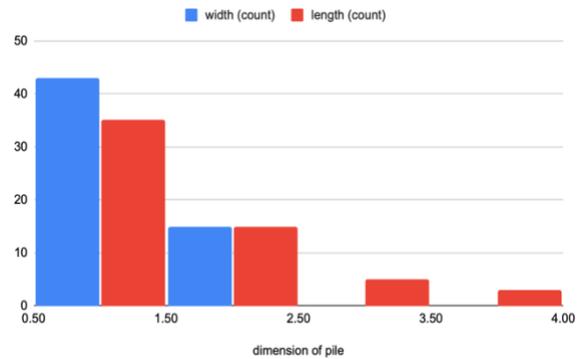

Figure 9 Dimension of pile for training

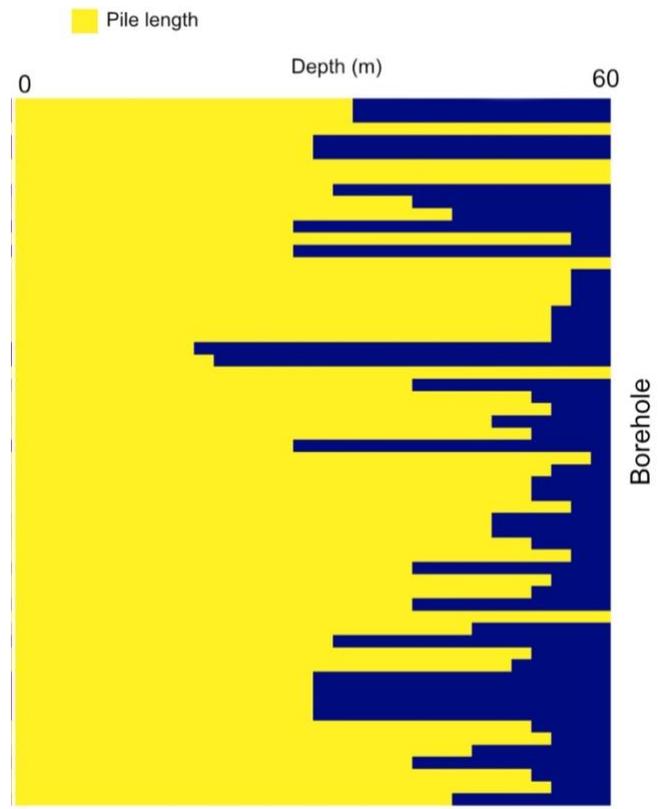

Figure 10 Pile length in soil layer

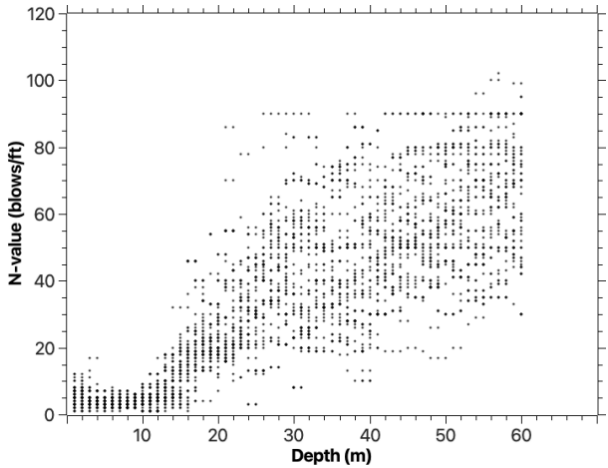

Figure 11 Value of SPT with depth

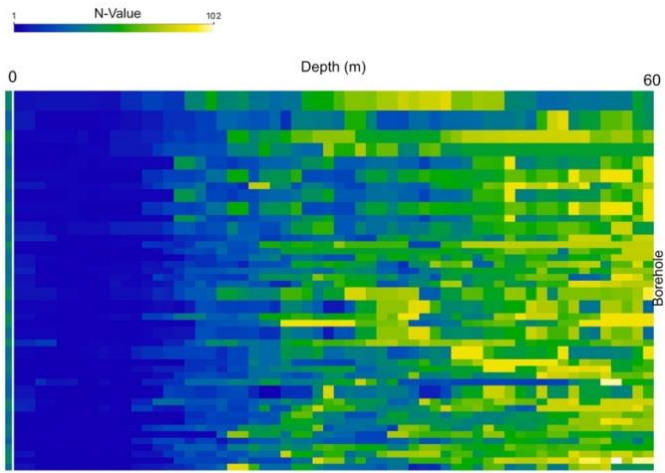

Figure 12 Value of SPT VS depth of each borehole

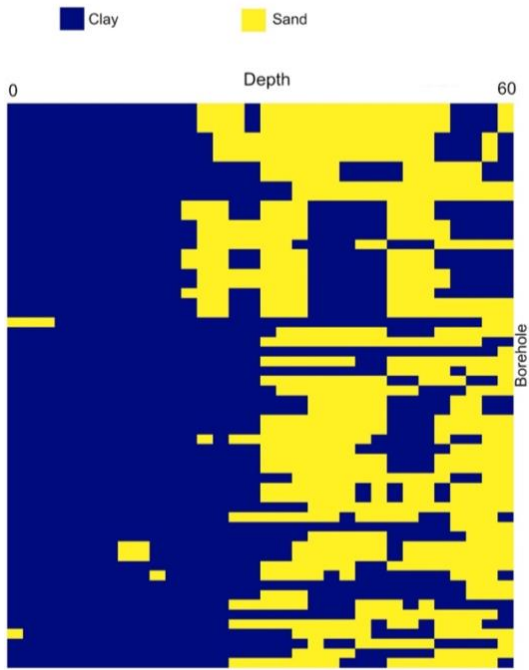

Figure 13 The location of clay and sand in each borehole

his study investigated the variation of SPT-N value with depth using soil profile data, which were encoded and used to predict the load deformation curve of pile. The undrained shear strength from unconfined compression tests was converted to SPT-N value using Equation 2 (see previous section). Figures 11 and 12 illustrates the distribution of SPT-N value with depth for all boreholes subjected to pile loading test. The SPT-N value varied slightly at shallow depths, indicating a uniform layer of very soft clay. However, the SPT-N value increased considerably after 18 m depth, where a sand layer was present (see Figure 13). The sand layer exhibited a high variation in thickness and properties, as well as interbedded clay layers with higher strength than the upper clay layer. The soil profile comprised alternating layers of clay and sand up to 60 m depth. A thin layer of sand at the ground surface represented the fill layer at the site. The trained data showed high variation in soil profile, which was influenced by the results of the Standard Penetration Test (SPT). The SPT was a field test that depended on various factors, such as the operator, the energy loss due to the machine, and the borehole condition. Therefore, finding a pattern for the data with high variation was very challenging for a deep learning model. We opted for an attention mechanism over other deep learning architectures, such as Long Short-Term Memory (LSTM) or Convolutional Neural Network (CNN), because we wanted to give attention to all tokens for each depth of soil. The tokens for each depth of soil were correlated with the other soil layers. However, this method required high computational resources for training and applying the model (generative AI).

## 6. MODEL TRAINING

Data normalization was performed before sending the vector to train a transformer. Data normalization can make the features have a similar scale, avoid activation function saturation, and prevent numerical errors. It can help the model learn faster and better. The pile feature vector was normalized with the min-max scaler concept, which preserves the shape of the original distribution, but does not reduce the effect of outliers in each type of feature. The following equation was used for min-max scaling:

$$x = \frac{X - X_{min}}{X_{max} - X_{min}} \qquad (7)$$

The SPT-N value and loading sequence of pile were scaled by dividing by a constant. The normalized data was obtained by the following equation:

$$x_{scaled} = x/F \qquad (8)$$

The factor, $F$, for SPT-N value and loading of pile was set to be 50 and 40000, respectively.

We tuned the hyperparameters of the transformer model to achieve the best convergence and stability, as shown in Table 2. However, the model complexity, such as the number of heads and layers for attention, led to model collapse or instability. Model collapse refers to the failure of the model to capture the diversity of the data distribution and generate only a few modes of outputs. This can result from overfitting, or from a poorly designed or misaligned objective function [27]. We set the encoder head size to 10, based on the input encoding token dimension of 7. We used a smaller decoder head size of 3, due to the lower dimension of the output shift right for the encoder. We applied a dropout layer of 0.1 to prevent overfitting. We used 10 loops for attention, which was the maximum possible without causing model collapse at high epochs. We modified the multilayer perceptron (MLP) by using five layers of conventional neural network with sigmoid activation function, and a linear activation function for the last layer as a regression task. This was different from previous large language models (LLM) that used cross-entropy or Softmax activation function for classification. We also added a dropout layer for regularization. The total number of trained parameters for this model was 547,321, which was not high compared to other sequential models.

Table 2 hyperparameters of model

| Layer | | Detail |
|---|---|---|
| Encoder | Multi head attention | head size=10, number of head =60, dropout =0.1, loop =10 |
| Decoder | Multi head attention | head size=3, number of head =60, dropout =0.1, loop =10 |
| | Multilayer perceptron | Dropout(0.1), 200, 100, 50, 5,1 |
| Train Parameters | | 547,321 parameters |

Table 3 The trained results of transformer model

| | MSE x$10^{-4}$ | MAPE (%) | MAE |
|---|---|---|---|
| Train | 2.50 | 5.84 | 0.0123 |
| Validation | 6.28 | 7.64 | 0.0183 |
| Test | 5.89 | 5.72 | 0.0185 |

We used the Notable GPU [28], a cloud-based server for running Python scripts, to train a transformer model with mean square error (MSE) as the loss function. The training data consisted of 90% of the original data, while the remaining 10% was split equally into validation and test sets. We set the batch size to 450 and trained the model for 4000 epochs, each taking 2 seconds. Figure 14 shows the training and testing losses as a function of epochs. The model selection criterion was the lowest MSE achieved during training. Both losses converged to a similar value of approximately $2\times10^{-4}$, indicating low overfitting. The model also performed well on both training and testing data, achieving a mean absolute percentage error (MAPE) of around 5%, which confirmed its generalization ability (Table 3). The mean absolute error (MAE) and MSE were also low and of similar magnitude for both data sets.

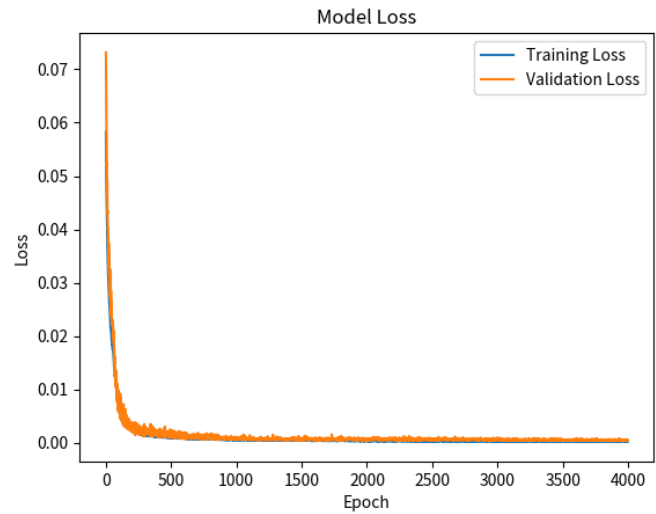

a) Loss (MSE) of model during training

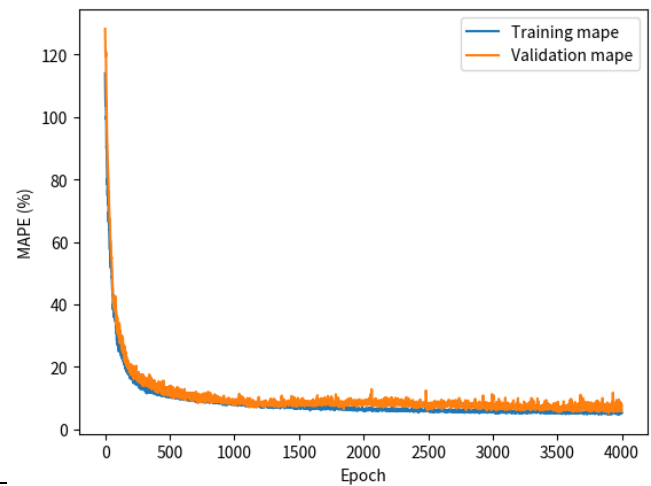

b) MAPE of model during training

Figure 14 The Loss and MAPE during training

## 7. MODEL PREDICTION

We used the trained transformer model to predict the load-deformation curve of 58 pile loading tests. We initialized the curve with a zero vector for all the sequences and fed it into the transformer along with the soil profile token and feature vector (Fig. 15). The model then predicted the next data point in the sequence. We appended the prediction to the predicted sequence and repeated the process until we reached a predefined number of iterations. This way, we generated the load-deformation curve of each pile. The predicted and actual load deformation curves of pile were compared in Fig. 16. The transformer model showed a satisfactory performance with an excellent fit to the actual results under different soil profile conditions and pile lengths. This trained model can be used to predict the load deformation curve of pile in Bangkok Subsoil. It can also be used as a tool for selecting the type and length of pile that can achieve the desired behavior of pile.

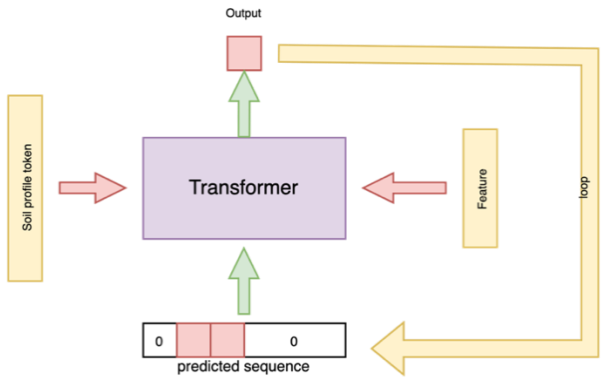

Figure 15 Model prediction

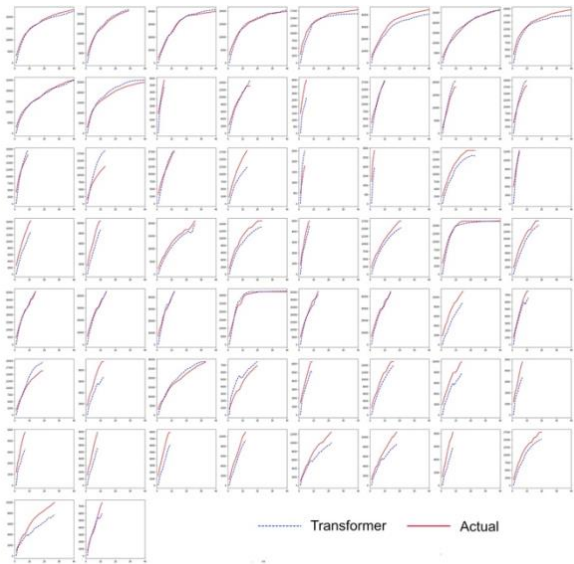

Figure. 16 The prediction of load deformation of pile from transformer model

the encoder and one for the feature vector of the decoder. The results showed that the stiffness of the load-deformation curve increased with increasing pile length (Fig. 19).

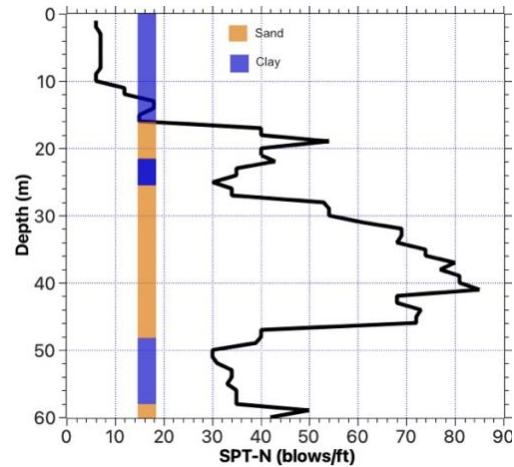

Figure 17 Soil Profile for Parametric study

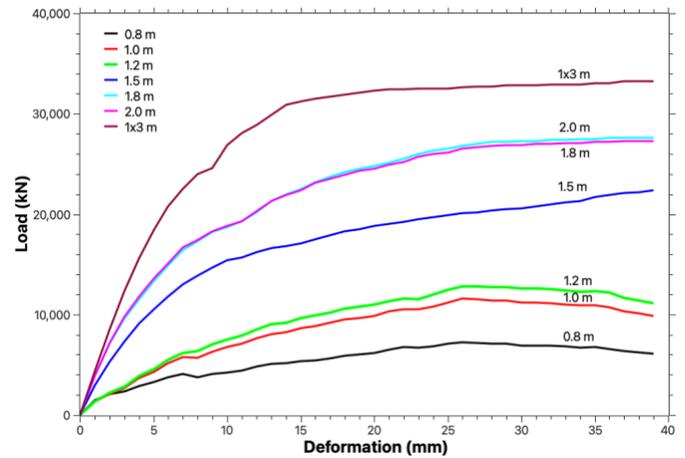

Figure 18 The load deformation curve of pile with different pile diameter

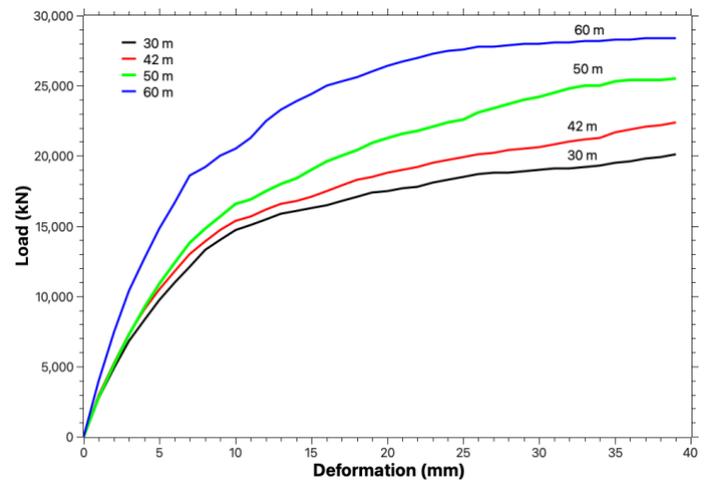

Figure 19 The load deformation curve of pile with different pile length

## 8. PARAMETRIC STUDY

A parametric study was conducted to evaluate the performance of the trained transformer model by varying the pile cross section, pile type, and pile length. The soil profile consisted of alternating layers of clay and sand with different values of standard penetration test (SPT-N) (Fig. 17). The soil profile was encoded as a token for each layer with the value of SPT-N and the soil type (clay or sand). This token was used as the input for the trained transformer. The pile length tip was initially set at 42 m on a sand layer. The load-deformation curve of the pile with varying pile size is shown in Fig. 18. The results showed that the trained model captured the effect of pile size on the stiffness and capacity of the pile. The rectangular pile or barrette had the highest load capacity and tended to converge at about 32,200 kN. The circular pile with a diameter of 0.8 m had the lowest load capacity. The effect of pile length on the load-deformation curve was also investigated using a circular pile with a diameter of 1.5 m. The pile length varied from 30 to 60 m. The pile length was encoded as two input vectors: one for

The proposed transformer-based model architecture showed excellent performance for detecting patterns from the load-deformation curves of pile load tests. The trained model is available for download from the authors' GitHub repository in the appendix. This model can be used as a pre-trained model and fine-tuned for different pile types or soil conditions. In the future, more pile load test data in the Bangkok area could be used to fine-tune this model to improve its ability to predict the pile behavior under different circumstances than this research. The proposed framework can be applied to other geotechnical problems for predicting the behavior of geotechnical systems. The soil profile and features can be encoded and sent to the decoder to predict the behavior of geotechnical problems such as wall movement, bearing capacity, or excavation problem. However, the limitation of the transformer is the requirement of high computational resources to run or train the model. It requires a computer with a graphics processing unit (GPU) and high memory.

## 9. CONCLUSION

This paper introduces a novel deep learning model based on the transformer architecture to predict the load-deformation behavior of bored piles in Bangkok subsoil1. The main contributions and findings of this paper are:
- The proposed model uses a comprehensive database of geotechnical properties and pile load tests in Bangkok to encode the soil profile and pile features as input, and generate the load-deformation curve as output. The model also incorporates the previous sequential data of load-deformation curve into the decoder, similar to the language translation model.
- The model shows a satisfactory accuracy and generalization ability, as well as insights into the advantages of the transformer-based approach. The model also demonstrates a low level of overfitting during the training stage. The mean absolute percentage error of test data could be low as 5.78 %. The model can generate the load-deformation curve of 58 cases of pile load test with an impressive performance.
- The model can also be used for parametric analysis and design optimization of piles under different soil and pile conditions. The trained model is publicly available on GitHub for geotechnical engineers to use and fine-tune.
- This paper contributes to the field of geotechnical engineering by providing a valuable tool for accurately predicting pile behaviors in both Bangkok's subsoil and other regions. The proposed framework can be applied to other geotechnical problems for predicting the behavior of geotechnical systems. The soil profile and features can be encoded and sent to the decoder to predict the behavior of geotechnical problems such as wall movement, bearing capacity, or excavation problem.
- 

## APPENDIX

The GitHub repository [https://github.com/Sompote/transformer_pile] contains the trained model and an example of generating the load deformation curve of pile.

## LIST OF SYMBOL

| Symbol | Description |
|---|---|
| $Q$ | Query vector |
| $V$ | Value vector |
| $K$ | Key vector |
| $W^o$ | Learn linear transformation vector |
| $W_i^Q$ | Learn linear transformation vector for query |
| $W_i^K$ | Learn linear transformation vector for key |
| $W_i^V$ | Learn linear transformation vector for value |
| $N$ | Standard penetration value |
| $S_u$ | Undrained shear strength |
| $x_{normalized}$ | Normalized value |
| $X$ | Value of data |
| $\mu$ | Mean value |
| $\sigma$ | Standard deviation |
| $\epsilon$ | Constant |
| $q$ | query |
| $k$ | key |
| $v$ | value |
| $W_i^Q$ | weight vector for query |
| $W_i^K$ | weight vector for key |
| $W_i^V$ | weight vector for value |
| $d_k$ | scaling factor |
| $x_{scaled}$ | Scaler data |
| $x_{min}$ | minimum value of data |
| $x_{max}$ | maximum value of data |
| $x$ | data |
| $F$ | factor |